\documentclass{article}

\usepackage{PRIMEarxiv}
\usepackage[utf8]{inputenc} 
\usepackage[T1]{fontenc}    
\usepackage{hyperref}       
\usepackage{url}            
\usepackage{booktabs}       
\usepackage{amsfonts}       
\usepackage{nicefrac}       
\usepackage{microtype}      
\usepackage{lipsum}
\usepackage{fancyhdr}       
\usepackage{graphicx}       
\graphicspath{{media/}}     
\usepackage{natbib}

\usepackage{graphicx}
\usepackage{makecell} 
\usepackage{multirow} 

\usepackage{algorithm}
\usepackage{algpseudocode}

\usepackage[most]{tcolorbox}
\usepackage[colorinlistoftodos]{todonotes}

\begin{document}
\pagestyle{fancy}
\thispagestyle{empty}
\rhead{ \textit{ }} 

\fancyhead[LO]{Fine-Tuning vs. Distillation for LLM Compression}

\title{Constrained Edge AI Deployment: Fine-Tuning vs. Distillation for LLM Compression}

\author{
  Jacob Sander$^{1}$, David Moe$^{2}$, Achraf Cohen$^{2}$, Brent Venable$^{3,4}$, Venkat R. Dasari$^{5}$, Brian Jalaian$^{1,3,4}$ \\
  $^{1}$Department of Computer Science, University of West Florida, Pensacola, USA \\
  $^{2}$Department of Mathematics and Statistics, University of West Florida, Pensacola, USA \\
  $^{3}$Department of Intelligent Systems and Robotics, University of West Florida, Pensacola, USA \\
  $^{4}$Institute of Human and Machine Cognition, Pensacola, USA \\
  $^{5}$DEVCOM Army Research Laboratory, Aberdeen Proving Ground, USA
}
\maketitle

\begin{abstract}
Modern foundational models are often compressed via a combination of structured pruning and re-training to meet the strict compute, memory, and connectivity constraints of edge deployments. 
While state-of-the-art pruning schemes target the entire Transformer, we adopt a simple, layer-wise $L_2$‐norm pruning on only the MLP blocks as a fixed baseline. Our focus is not on achieving maximal compression, but on isolating the impact of the re-training loss function: \textbf{(i) Fine-tuning} with Cross-Entropy (L2PFT), which requires labeled data, versus \textbf{(ii) Self-Distillation} with KL‐divergence, which leverages only teacher logits (no labels) (L2PSD). We evaluate both pipelines on the OLMo2-7B-SFT model for CommonsenseQA suitable for intermittent or denied connectivity scenarios typical of edge networks. Under identical pruning schedules, KL-based distillation matches or exceeds CE fine-tuning in test accuracy, demonstrating that, even with a basic MLP‐only pruning, the choice of loss function materially affects compressed model recovery in resource-constrained environments.  
\end{abstract}


\section{Introduction}

Recent advances in artificial intelligence (AI) have transformed a range of fields, from computer vision and autonomous navigation to military applications and healthcare. The capabilities and complexity of AI models are rapidly increasing, fueled by improvements in both hardware and algorithms \cite{Lin2022}. However, deploying these large foundational models on resource-constrained edge computing platforms can be challenging due to their high computational intensity, limited on-device power and memory, and intermittent or denied connectivity common in tactical-edge operations.

Typical optimization pipelines for edge AI combine structured pruning, knowledge distillation, and low-rank approximation to reduce model size and computation. Pruning removes or deactivates neurons or channels, quantization lowers weight precision, and neural architecture search redistributes capacity \cite{white2023neural}. Collaborative edge frameworks such as EdgeShard further optimize model partitioning across devices \cite{ZhangEdgeShard}, and comprehensive surveys cover model compression and edge AI strategies in depth \cite{sander2025accelerating, Singh2023EdgeAA}.

Knowledge distillation \cite{hinton}, particularly self-distillation when teacher and student share the same architecture, transfers soft targets from a larger model to a smaller one. This approach has been shown to compress transformers effectively while preserving performance \cite{Chen2024, ONeill2023}. Combining pruning with distillation often yields better trade-offs between model size, accuracy, and inference speed than either technique alone \cite{Than2024, Sreenivas2024LLM}. However, it remains unclear whether post-prune recovery gains arise primarily from the pruning strategy or from the choice of loss function used during re-training.

In this work, we isolate and examine the impact of two re-training losses under a fixed, MLP-only pruning baseline. Specifically, we apply layer-wise $L_2$-norm pruning exclusively to the MLP layers of the Transformer and then compare:
\begin{itemize}
  \item \textbf{L2PFT}: Cross-Entropy(CE) fine-tuning (label-dependent)
  \item \textbf{L2PSD}: KL-divergence self-distillation (label-free)
\end{itemize}
This controlled setup allows us to ask: \emph{Does post-prune performance depend more on the pruning mechanism or on the loss function used for recovery?} We evaluate both pipelines on the OLMo2-7B-SFT model for CommonsenseQA, demonstrating that KL-based distillation matches or outperforms CE fine-tuning under identical compression schedules. Our findings highlight a non-trivial role for the loss function in compressed-model optimization, with implications for designing efficient, data-sparse edge AI systems.

The rest of this paper is organized as follows. Section \ref{relatedwork} reviews related work on pruning and distillation. Section \ref{methods} describes our methodology. Section \ref{results} presents the experimental results and discussion, and Section \ref{conclusion} concludes the paper.

\section{Background and Related Work}
\label{relatedwork}
The goal of any AI model is to approximate an unknown function $f$; in particular, for supervised learning, we aim to approximate $f: \mathbb{R}^d \rightarrow \mathbb{R}^m$ using data $\mathcal{D} = \{ x_i, f(x_i)\}$ where $i= 1, \ldots, N$. This is achieved by selecting different models whose parameters $\theta$ are learned using the training data, which involves minimizing a cost function, $\mathcal{L(\theta)}$, chosen by the user.

Foundational models are computationally intensive and highly over-parameterized, making model compression techniques such as pruning and knowledge distillation critical for efficiency. The most effective techniques balance sparsity, performance retention, and hardware compatibility.

In optimization, when training a neural network, we aim to find parameters that minimize the loss function. The loss landscape can be complex, with many local minima. But in over-parameterized networks (like foundational models), it's shown that there are many global minima—points where the loss is as low as possible. \cite{chaudhari2018stochastic} discusses flat minima and the geometric structure of the loss. \cite{sagun2017empirical} analyzes the Hessian spectrum, revealing flat directions in the loss landscape. Distillation strategies impact the geometric properties of neural network loss landscapes in measurable ways. \cite{pham2022revisitingselfdistillation,sarn2023} 
Self-distillation typically results in flatter minima, which is linked to improved generalization, a result of self-distillation's regularization effect on model parameters. \cite{Kim2021} \cite{mobahi2020self}


Pruning is a technique used to reduce the size of neural network models by removing specific rows from one linear layer and corresponding columns in the next layer. This decreases the model's size and eliminates certain terms from the intermediate activation vector. The concept was first explored by LeCun et al. \cite{lecun1989optimal} in their paper on Optimal Brain Damage, using second-order derivatives to assess the importance of different components. The modern taxonomy of pruning is rich. Researchers have developed numerous approaches to pruning, combining choices of timing (before, during, or after training), iterative or one-shot implementation, criterion (the metric to decide which elements to delete), and structures targeted (embedding channel, individual weights, entire layers, etc). \cite{sander2025accelerating} Structured pruning, which selectively removes less critical components of a neural network, has emerged as a promising method for improving LLM efficiency. \cite{Than2024}



LLMs have become a cornerstone of modern natural language processing (NLP), enabling significant advancements in text generation, reasoning, and multimodal applications. The foundation of these models can be traced to architectures such as the Transformer \cite{vaswani2017attention}, which introduced self-attention as a scalable alternative to recurrent networks. Following this breakthrough, autoregressive pre-trained models like GPT-3 \cite{brown2020language} and instruction-tuned variants such as GPT-4 \cite{openai2023gpt4} have demonstrated remarkable capabilities across diverse language tasks.

Parallel to these efforts, open-weight models like LLaMA \cite{touvron2023llama} have pushed the boundaries of accessibility and efficiency, offering viable alternatives to proprietary models. These works emphasize parameter-efficient training, adaptation to multilingual corpora, and improvements in controllability and alignment.


OLMo (Open Language Model) is part of a growing movement toward open-source, transparent LLMs. Developed to serve as an accessible alternative to proprietary models, the initial release of OLMo provided public access to training data, logs, and evaluation benchmarks, fostering greater reproducibility in large-scale language modeling.\cite{olmo20242olmo2furious}

OLMo2 builds on this foundation, refining both its training methodology and architecture. The model achieves improved generalization across various NLP tasks by incorporating a more diverse and higher-quality dataset. \cite{olmo20242olmo2furious} Architectural modifications, including optimization in attention mechanisms and tokenization strategies, enhance efficiency and performance. Unlike earlier models that rely on instruction tuning as a post-processing step, OLMo2 integrates adaptive alignment techniques throughout pretraining, improving its usability in downstream applications. Designed for scalability, OLMo2 is released with multiple parameter sizes and public benchmarks, enabling broader community engagement in fine-tuning and evaluation. \cite{olmo20242olmo2furious}


\vspace{-1em}
\section{Methodology}
\label{methods}
The proposed framework for model compression involves structured pruning, quantization, and self-distillation to enhance inference efficiency while maintaining accuracy.

\subsection{Pruning}

To achieve efficient model compression, our framework employs structured pruning targeting the transformer's multilayer perceptron (MLP) layers. The MLP layers account for the majority of the model’s parameters, and their structured pruning enables significant computational speedups without requiring sparsity-aware kernels. Our investigation into self-distillation, known for its regularizing effect on model parameters,  \cite{mobahi2020self} guided the design of our pruning strategy. Specifically, we hypothesized that self-distillation’s ability to regularize weight magnitudes could enhance the effectiveness of weight-based pruning criteria. Motivated by this, we selected the L2 norm of weight magnitudes in the MLP layers as our pruning criterion. Mathematically, consider $w^b_{(i,j)}$ is the transformer's weights matrix elements of block $b$, then for each row $i$ and block $b$:
\begin{equation}
\label{l2}
    \|W^b_{(i)}\|_2 = \sqrt{\sum_{j} (W^b_{(i,j)})^2}
\end{equation}

The L2 norm offers a data-free evaluation method, enabling efficient neuron selection without the computational overhead of dataset-dependent metrics. By leveraging self-distillation’s regularization, we aim to improve the quality of the L2 norm criterion, as regularized weight magnitudes should better reflect parameter importance. This approach allows us to evaluate whether self-distillation’s regularizing effect enhances the performance of structured pruning compared to fine-tuning. If our hypothesis holds, self-distillation should lead to improved pruning decisions, resulting in higher model accuracy at equivalent sparsity levels. We evaluate this criterion across each row of the UP ($W_u)$ and GATE ($W_g$) layers in each attention block, then prune the rows of UP and GATE and the corresponding columns of the DOWN ($W_d$) layer. Our proposed Algorithm is presented in Algorithm \ref{alg:structured_pruning}.

After pruning, we restore the accuracy of the model either using conventional fine-tuning, or self-distillation, as follows:

\begin{algorithm}
\caption{Structured Pruning}

\begin{algorithmic}[1]
\Require \textbf{k} - percent of neurons to prune
\linespread{0.5}
    \State \textbf{Prune linear layers:}
\State Compute L2 norm of GATE \( W_g \) and UP layers \( W_u \) following  \eqref{l2}
\State Compute average L2 norm for each row $i$ and block $b$:
\[
W^b_{(i)} =(\|W^b_{(i)g}\|_2 + \|W^b_{(i)u}\|_2)/2
\]
\State Create a mask on the average L2 norm:
\[
\mathbf{1}_k(w^b) = \mathbb{I}(w^b > P_k(w^b)); \quad w^b=\{W^b_{(1)}, \ldots\},
\]
where $\mathbb{I}$ is the element-wise indicator function. $P_k(w^b)$ is the $k$-th percentile of $w^b$.
\State Prune GATE (g), UP (u) rows, and DOWN (d) columns:
\[
W^{b}_{l,\text{pruned}} = W^b_l \odot \mathbf{1}_k(W^b_l) \text{ and }  W^{b}_{d,\text{pruned}} = W^b_d \odot \mathbf{1}_k(W^b_d)^T,
\]
where $l \in \{g, u\}$ and $\odot$ denotes the Hadamard product

\end{algorithmic}
\label{alg:structured_pruning}
\end{algorithm}


\subsection{Fine-Tuning}
To recover the accuracy of the pruned model, we apply conventional fine-tuning, as detailed in Algorithm \ref{alg:prune_finetune}. The pruned model is optimized using the cross-entropy loss with respect to the ground-truth labels, defined as: 

\begin{equation}
    \mathcal{L}_{\text{CE}}(y_b, p_S) = - \sum_{i} y_b(i) \log p_S(i), \label{eq:ce_loss}
\end{equation}

where $y_b(i)$ represents the one-hot encoded ground-truth label for class $i$, and $p_S(i)$ is the student model’s predicted probability for class $i$. Fine-tuning adjusts the remaining weights in the pruned UP, GATE, and DOWN layers to minimize $\mathcal{L}_{\text{CE}}$, using stochastic gradient descent over the training dataset. Fine tuning directly optimizes the model for task-specific performance, serving as a baseline to evaluate the benefits of self-distillation’s regularization in the pruning process. By comparing fine-tuning to self-distillation, we assess whether the latter’s soft targets and regularizing effects yield superior accuracy at equivalent sparsity levels.

\begin{algorithm}
\caption{L2PFT: Joint Pruning \& Iterative Fine-tuning }
\begin{algorithmic}[1]
\Require Dataset \(D\) (multiple-choice questions and single-token answers)
\Require Pretrained model \(F\) (FP32), pruning rate \(k\), number of epochs \(n\)
\linespread{0.9}
\small
\State \textbf{Quantize model:} Convert \(F\) from FP32 to FP16

\For{\( \text{epoch} = 1 \) to \(n\)}
    \State \textbf{Structured Pruning (\(F\))}
    \State \textbf{Fine Tune:}
    \For{each batch \(b \in D\)}
        \State Compute Cross-Entropy loss as in \eqref{eq:ce_loss}

   j     \State Perform backward pass and update model parameters:
        \[
        \theta_S \leftarrow \theta_S - \eta \nabla_{\theta_S} \mathcal{L}_{\text{CE}}
        \]
    \EndFor
\EndFor
\end{algorithmic}
\label{alg:prune_finetune}
\end{algorithm}
\subsection{Self-Distillation}

To leverage the regularizing effects of self-distillation for post-pruning accuracy recovery, we apply self-distillation to the pruned model, as detailed in Algorithm \ref{alg:prune_distill}. In this setup, the unpruned model serves as the teacher, providing soft targets to guide the training of the pruned student model. To optimize computational efficiency and reduce GPU utilization, teacher logits are pre-computed before training and stored as soft targets in an augmented dataset, paired with the corresponding questions from the CommonsenseQA dataset.

The softmax probabilities for the teacher and student models with temperature scaling are defined as:
\begin{align}
p_T(i) &= \frac{\exp(t_i / T)}{\sum_j \exp(t_j / T)}, \label{eq:teacher_softmax} \\
p_S(i) &= \frac{\exp(s_i / T)}{\sum_j \exp(s_j / T)}, \label{eq:student_softmax}
\end{align}
where \( t_i \) and \( s_i \) are the logits of the teacher and student models, respectively, and \( T \) is the temperature parameter.

The KL divergence loss for self-distillation is given by:
\begin{equation}
\mathcal{L}_{\text{KL}}(p_T, p_S) = T^2 \sum_i p_T(i) \log \frac{p_T(i)}{p_S(i)}, \label{eq:kl_loss}
\end{equation}
where the \( T^2 \) factor scales the loss to account for the temperature-adjusted targets. This loss is minimized using stochastic gradient descent, adjusting the remaining weights in the pruned UP, GATE, and DOWN layers. By training with soft targets, self-distillation regularizes the weights, potentially improving the L2-norm-based pruning criterion’s effectiveness compared to fine-tuning’s hard-label optimization. This approach allows us to test our hypothesis that self-distillation’s regularization leads to better pruning decisions and higher model accuracy at equivalent sparsity levels.

\begin{algorithm}
\caption{L2PSD: Joint Pruning \& Self-Distillation}
\begin{algorithmic}[1]
\Require Dataset \(D\) (multiple-choice questions and single-token answers)
\Require Pretrained model \(F\) (FP32), pruning rate \(k\), number of epochs \(n\)
\linespread{0.9}
\small
\State \textbf{Obtain teacher logits:}
    \For{each batch \(b \in D\)}
        \State Generate output logits using \(F\)
        \State Append teacher logits to \(D\) to create  \(D'\)
    \EndFor
        
\State \textbf{Quantize model:} Convert \(F\) from FP32 to FP16

\For{\( \text{iteration} = 1 \) to \(n\)}

    \State \textbf{Structured Pruning} (\(F\))

    \State \textbf{Distill knowledge:}
    \For{each batch \(b \in D'\)}
\State Extract teacher logits \( \mathbf{t}_b \) from \( D' \): ${t}_b = f_T(x_b)$
        \State Compute student logits \( \mathbf{s}_b \) from \( F \): ${s}_b = f_S(x_b)$

        \State Compute targets $p_T$ and $p_S$ as in \eqref{eq:teacher_softmax} and \eqref{eq:student_softmax}:

        \State Compute KL divergence loss following Equation \eqref{eq:kl_loss}:
        

        \State Perform backward pass and update student parameters:
        \[
        \theta_S \leftarrow \theta_S - \eta \nabla_{\theta_S} \mathcal{L}_{\text{KL}}
        \]
    \EndFor
\EndFor
\end{algorithmic}
\label{alg:prune_distill}
\end{algorithm}
%

\section{Experiments and Results}
\label{experiments}
We evaluate our model compression algorithms on the task of multiple-choice question answering using the CommonsenseQA dataset, \cite{talmor-etal-2019-commonsenseqa}, which tests commonsense reasoning through questions with five answer choices. The dataset has 12,102 questions split into 9,741 training, 1,221 validation, and 1,140 test multiple-choice questions.
We use a two-shot prompting strategy to provide context and improve model performance, as this approach has been shown to enhance reasoning in similar tasks \cite{multishot_brown}. We assess model performance using the following metrics: 
\begin{itemize}
    \item Accuracy on the hold-out test set to evaluate generalization
    \item Accuracy on the training set to monitor overfitting
    \item Inference FLOPs to quantify computational efficiency. Estimated \textit{a priori} from model architecture using package ptflops \cite{ptflops}
    \item Wall-clock time (measured in seconds) for a single forward pass during inference to measure real-world latency
    \item Shannon entropy of the output distribution over answer tokens to gauge prediction uncertainty
\end{itemize}

We conducted experiments using the pre-trained OLMo2-7B-SFT model, \cite{olmo20242olmo2furious}, a 7-billion-parameter model fine-tuned for instruction-following, selected for its balance of performance, efficiency, and openness. We conducted a data ablation study to evaluate the robustness of our pruning strategies under varying data conditions. In addition to training on the full dataset, we subsampled the training data to 75\%, 50\%, and 25\% of its original size and evaluated all of the metrics at these levels. Both pruning approaches start with an unpruned OLMo2 model, loaded using a standardized model initialization function. The pruned models are fine-tuned on the training dataset and evaluated on a held-out test set on the five metrics above. The model is implemented in PyTorch using the HuggingFace Transformers library and parallelized using HuggingFace Accelerator. Experiments were completed on 8 A40 GPUs. Our key findings are as follows:
\vspace{-.25em}
\label{results}

\begin{table}[htbp]
    \centering
    \resizebox{\columnwidth}{!}{
    \begin{tabular}{c|c|c|c|c|c|c|c|c|c|c}
        \toprule
        \multirow{2}{*}{\textbf{Retention}} & 
        \multicolumn{2}{c|}{\textbf{Train Accuracy}} & 
        \multicolumn{2}{c|}{\textbf{Test Accuracy}} & 
        \multicolumn{2}{c|}{\textbf{Entropy}} & 
        \multicolumn{2}{c|}{\textbf{Inference Time (s)}} & 
        \multicolumn{2}{c}{\textbf{Inference FLOPs}} \\
        \cmidrule(lr){2-3} \cmidrule(lr){4-5} \cmidrule(lr){6-7} \cmidrule(lr){8-9} \cmidrule(lr){10-11}
        & \makecell[c]{Fine-\\Tune} & \makecell[c]{Distill} 
        & \makecell[c]{Fine-\\Tune} & \makecell[c]{Distill} 
        & \makecell[c]{Fine-\\Tune} & \makecell[c]{Distill} 
        & \makecell[c]{Fine-\\Tune} & \makecell[c]{Distill} 
        & \makecell[c]{Fine-\\Tune} & \makecell[c]{Distill} \\
        \midrule
100\% & 0.717 & 0.717 & 0.747 & 0.747 & 0.537 & 0.537 & 2.789 & 2.789 & 1.570e+12 & 1.570e+12 \\
        91\% & 0.924 & 0.694 & 0.632 & 0.646 & 0.648 & 0.797 & 2.981 & 3.023 & 1.424e+12 & 1.424e+12 \\
        81\% & 0.971 & 0.702 & 0.544 & 0.561 & 0.692 & 0.739 & 2.649 & 2.648 & 1.250e+12 & 1.250e+12 \\
        70\% & 0.938 & 0.677 & 0.426 & 0.464 & 0.839 & 0.803 & 2.318 & 2.320 & 1.074e+12 & 1.074e+12 \\
        61\% & 0.896 & 0.652 & 0.327 & 0.380 & 1.023 & 0.925 & 2.052 & 2.029 & 9.146e+11 & 9.146e+11 \\
        50\% & 0.645 & 0.489 & 0.234 & 0.267 & 1.802 & 1.414 & 1.846 & 1.870 & 7.410e+11 & 7.409e+11 \\
        \bottomrule
    \end{tabular}
    }
    \vspace{0.25em}
    \caption{Ablation Study Results - 25\% of the CQA dataset}
    \label{tab:ablation_study_25}
\end{table}
\AtEndEnvironment{table}{\vspace{-0.25em}}     

\AtBeginEnvironment{table}{\vspace{-0.25em}}
\begin{table}[htbp]
    \centering
    \resizebox{\columnwidth}{!}{
    \begin{tabular}{c|c|c|c|c|c|c|c|c|c|c}
        \toprule
        \multirow{2}{*}{\textbf{Retention}} & 
        \multicolumn{2}{c|}{\textbf{Train Accuracy}} & 
        \multicolumn{2}{c|}{\textbf{Test Accuracy}} & 
        \multicolumn{2}{c|}{\textbf{Entropy}} & 
        \multicolumn{2}{c|}{\textbf{Inference Time (s)}} & 
        \multicolumn{2}{c}{\textbf{Inference FLOPs}} \\
        \cmidrule(lr){2-3} \cmidrule(lr){4-5} \cmidrule(lr){6-7} \cmidrule(lr){8-9} \cmidrule(lr){10-11}
        & \makecell[c]{Fine-\\Tune} & \makecell[c]{Distill} 
        & \makecell[c]{Fine-\\Tune} & \makecell[c]{Distill} 
        & \makecell[c]{Fine-\\Tune} & \makecell[c]{Distill} 
        & \makecell[c]{Fine-\\Tune} & \makecell[c]{Distill} 
        & \makecell[c]{Fine-\\Tune} & \makecell[c]{Distill} \\
        \midrule
        100\% & 0.717 & 0.717 & 0.747 & 0.747 & 0.537 & 0.537 & 2.789 & 2.789 & 1.570e+12 & 1.570e+12 \\
        91\% & 0.967 & 0.710 & 0.665 & 0.654 & 0.463 & 0.692 & 3.021 & 3.045 & 1.424e+12 & 1.424e+12 \\
        81\% & 0.985 & 0.711 & 0.593 & 0.594 & 0.437 & 0.663 & 2.678 & 2.672 & 1.250e+12 & 1.250e+12 \\
        70\% & 0.968 & 0.703 & 0.497 & 0.529 & 0.571 & 0.702 & 2.281 & 2.295 & 1.074e+12 & 1.074e+12 \\
        61\% & 0.934 & 0.685 & 0.414 & 0.464 & 0.792 & 0.819 & 2.036 & 2.056 & 9.146e+11 & 9.146e+11 \\
        50\% & 0.818 & 0.608 & 0.287 & 0.342 & 1.325 & 1.061 & 1.866 & 1.759 & 7.408e+11 & 7.408e+11 \\
        \bottomrule
    \end{tabular}
    }
    \vspace{0.25em}
    \caption{Ablation Study Results - 50\% of the CQA dataset}
    \label{tab:ablation_study_50}
\end{table}
\AtEndEnvironment{table}{\vspace{-0.25em}}

\AtBeginEnvironment{table}{\vspace{-0.25em}}
\begin{table}[htbp]
    \centering
    \resizebox{\columnwidth}{!}{
    \begin{tabular}{c|c|c|c|c|c|c|c|c|c|c}
        \toprule
        \multirow{2}{*}{\textbf{Retention}} & 
        \multicolumn{2}{c|}{\textbf{Train Accuracy}} & 
        \multicolumn{2}{c|}{\textbf{Test Accuracy}} & 
        \multicolumn{2}{c|}{\textbf{Entropy}} & 
        \multicolumn{2}{c|}{\textbf{Inference Time (s)}} & 
        \multicolumn{2}{c}{\textbf{Inference FLOPs}} \\
        \cmidrule(lr){2-3} \cmidrule(lr){4-5} \cmidrule(lr){6-7} \cmidrule(lr){8-9} \cmidrule(lr){10-11}
        & \makecell[c]{Fine-\\Tune} & \makecell[c]{Distill} 
        & \makecell[c]{Fine-\\Tune} & \makecell[c]{Distill} 
        & \makecell[c]{Fine-\\Tune} & \makecell[c]{Distill} 
        & \makecell[c]{Fine-\\Tune} & \makecell[c]{Distill} 
        & \makecell[c]{Fine-\\Tune} & \makecell[c]{Distill} \\
        \midrule
100\% & 0.717 & 0.717 & 0.747 & 0.747 & 0.537 & 0.537 & 2.789 & 2.789 & 1.570e+12 & 1.570e+12 \\
        91\% & 0.970 & 0.719 & 0.688 & 0.683 & 0.396 & 0.689 & 2.945 & 3.066 & 1.424e+12 & 1.424e+12 \\
        81\% & 0.990 & 0.720 & 0.606 & 0.615 & 0.403 & 0.680 & 2.714 & 2.670 & 1.250e+12 & 1.250e+12 \\
        70\% & 0.980 & 0.712 & 0.540 & 0.559 & 0.484 & 0.712 & 2.328 & 2.290 & 1.074e+12 & 1.074e+12 \\
        61\% & 0.960 & 0.703 & 0.452 & 0.486 & 0.656 & 0.747 & 2.038 & 4.309 & 9.146e+11 & 9.145e+11 \\
        50\% & 0.842 & 0.647 & 0.340 & 0.376 & 1.088 & 0.966 & 1.769 & 1.741 & 7.409e+11 & 7.408e+11 \\
        \bottomrule
    \end{tabular}
    }
    \vspace{0.25em}
    \caption{Ablation Study Results - 75\% of the CQA dataset}
    \label{tab:ablation_study_75}
\end{table}
\AtEndEnvironment{table}{\vspace{-0.25em}}

\AtBeginEnvironment{table}{\vspace{0em}}
\begin{table}[htbp]
    \centering
    \resizebox{\columnwidth}{!}{
    \begin{tabular}{c|c|c|c|c|c|c|c|c|c|c}
        \toprule
        \multirow{2}{*}{\textbf{Retention}} & 
        \multicolumn{2}{c|}{\textbf{Train Accuracy}} & 
        \multicolumn{2}{c|}{\textbf{Test Accuracy}} & 
        \multicolumn{2}{c|}{\textbf{Entropy}} & 
        \multicolumn{2}{c|}{\textbf{Inference Time (s)}} & 
        \multicolumn{2}{c}{\textbf{Inference FLOPs}} \\
        \cmidrule(lr){2-3} \cmidrule(lr){4-5} \cmidrule(lr){6-7} \cmidrule(lr){8-9} \cmidrule(lr){10-11}
        & \makecell[c]{Fine-\\Tune} & \makecell[c]{Distill} 
        & \makecell[c]{Fine-\\Tune} & \makecell[c]{Distill} 
        & \makecell[c]{Fine-\\Tune} & \makecell[c]{Distill} 
        & \makecell[c]{Fine-\\Tune} & \makecell[c]{Distill} 
        & \makecell[c]{Fine-\\Tune} & \makecell[c]{Distill} \\
        \midrule
100\% & 0.717 & 0.717 & 0.747 & 0.747 & 0.537 & 0.537 & 2.789 & 2.789 & 1.570e+12 & 1.570e+12 \\
        91\% & 0.981 & 0.721 & 0.705 & 0.697 & 0.366 & 0.638 & 2.979 & 2.978 & 1.424e+12 & 1.424e+12 \\
        81\% & 0.990 & 0.719 & 0.639 & 0.637 & 0.354 & 0.689 & 2.690 & 2.642 & 1.250e+12 & 1.250e+12 \\
        70\% & 0.987 & 0.717 & 0.579 & 0.598 & 0.423 & 0.680 & 2.310 & 2.329 & 1.074e+12 & 1.074e+12 \\
        61\% & 0.967 & 0.705 & 0.481 & 0.520 & 0.592 & 0.708 & 2.036 & 2.045 & 9.146e+11 & 9.145e+11 \\
        50\% & 0.898 & 0.657 & 0.365 & 0.390 & 1.005 & 0.874 & 1.740 & 1.841 & 7.408e+11 & 7.409e+11 \\
        \bottomrule
    \end{tabular}
    }
    \vspace{0.25em}
    \caption{Ablation Study Results - 100\% of the CQA dataset}
    \label{tab:ablation_study_100}
\end{table}
\AtEndEnvironment{table}{\vspace{-0.25em}}

\begin{figure}
    \centering
    \includegraphics[width=1\linewidth]{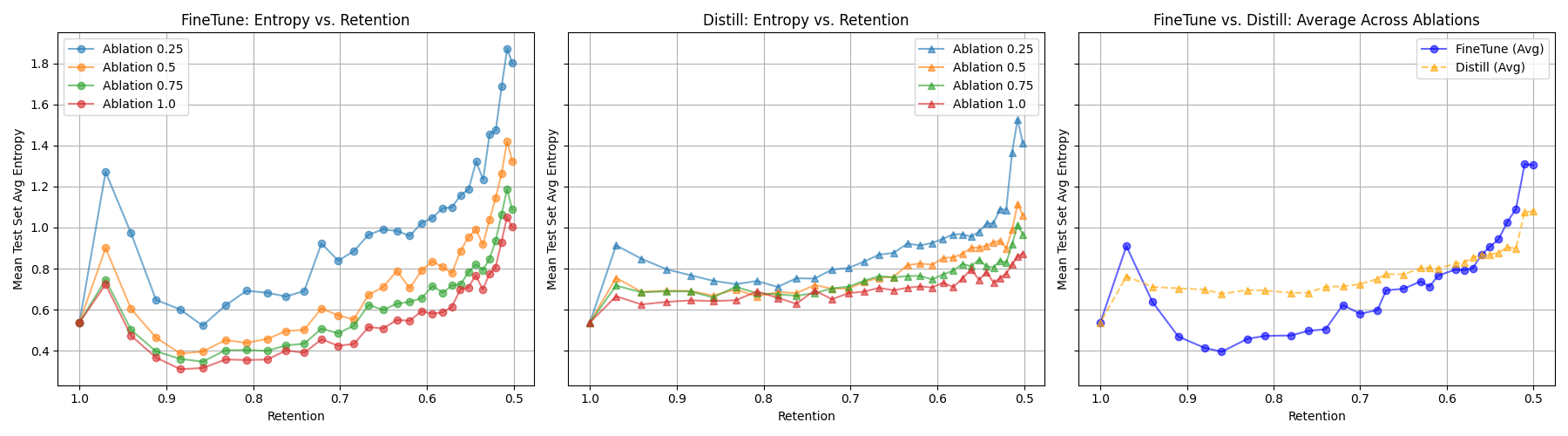}
    \vspace*{-2.5em}
    \caption{Entropy and Data Ablation; Distillation and Fine Tuning}
    \vspace{-1em}
    \label{fig:entropy_ablation}
\end{figure}

\begin{figure}
    \centering
    \includegraphics[width=1\linewidth]{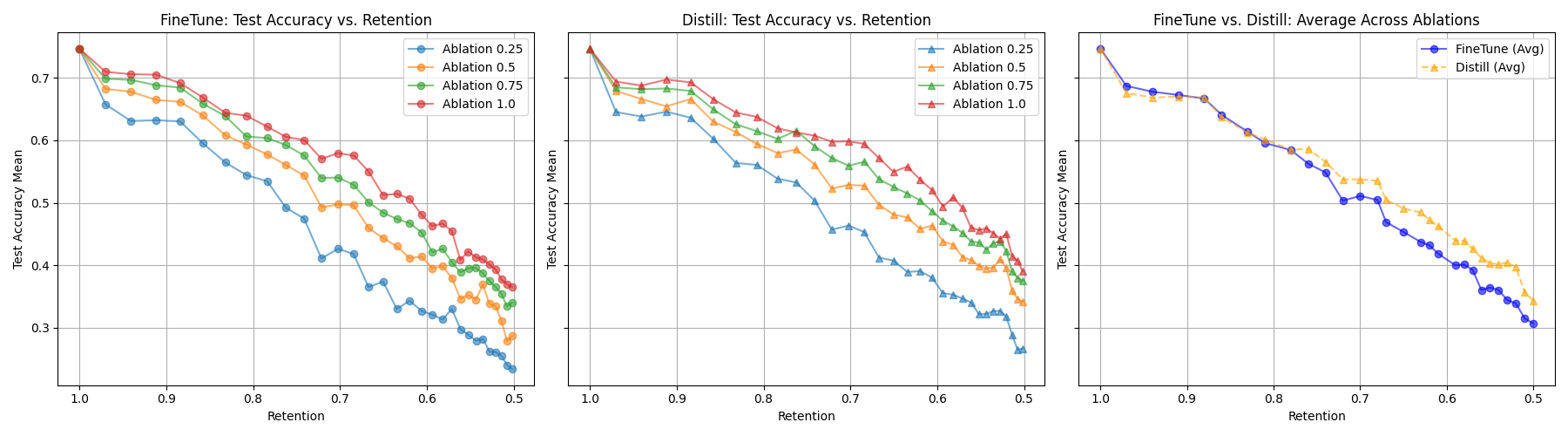}
    \vspace*{-2.5em}
    \caption{Test Accuracy and Data Ablation; Distillation and Fine Tuning}
    \vspace*{-1em}
    \label{fig:testacc_ablation}
\end{figure}

\vspace{.25em}




\vspace{-0.5em}
\section{Discussion}
Over the course of iterations, KL-divergence-based self-distillation ultimately outperforms CE-based fine-tuning on test set accuracy, generally by 3-5\% on test accuracy at 50\% retention.
This provides an answer to our question of whether dependence for post-pruning performance relies on the loss function used for recovery. The KL-based self-distillation matches and even outperforms the CE-based fine-tuning's ability to recover accuracy over successive iterations. 
In line with previous work on weight-magnitude-based neuron-structured pruning, we find that we begin experiencing accuracy decay after 90\% retention, in line with the perplexity-based results of magnitude pruning in \cite{Frantar_Alistarh_2023}. 
Increased data availability positively impacts accuracy uniformly across all tests, causing a 5-10 point accuracy increase between 25\% and 100\% ablation across the iterative pruning trajectory. See Fig. \ref{fig:testacc_ablation}.

At deployment, model uncertainty plays a vital role in safety-critical tactical environments. Conventional Uncertainty Quantification (UQ) methods depend on well-calibrated model output probabilities to flag incorrect responses. \cite{he2025surveyuncertaintyquantificationmethods} While our fine-tuned model does display comparable accuracy in the regime above 80\% retention, its entropy sharply decreases, creating the U-shaped graph in Fig.~\ref{fig:entropy_ablation}. By contrast, the self-distilled entropy remains flat until 60\% retention. This suggests that the model output probability distribution is much more concentrated in the fine-tuning method, and will be inappropriately confident in its responses. Such outputs will be more difficult to flag with model-confidence based UQ methods.
We also find that increased training data size uniformly decreased Shannon Entropy (see Fig. \ref{fig:entropy_ablation}), as the extended training shrinks the distribution of model predictions. This is a more appropriate decrease in entropy, as it is associated with increased accuracy.

We additionally note that in the tactical environment, labeled data can be expensive and rare; a method where a model labels its own dataset using self-distillation is a compelling and practical alternative to traditional supervised learning with a labeled dataset.
We acknowledge that these pruning results do not represent SoTA pruning-based compression. For example, NVIDIA's Minitron offers pruning-based compression for LLMs on the order of 50\% retention for accuracy decay of 2-4\%\cite{Sreenivas2024LLM}.
It does this by pruning along multiple axes, such as depth and embedding, not just the MLP layers of LLMs; by the time our pruning method reaches 50\% retention, only approximately 10\% of the original MLP neurons remain.


However, this study investigates whether post-prune performance depends more on pruning mechanisms or loss functions, demonstrating that KL-divergence self-distillation consistently matches or exceeds cross-entropy (CE) fine-tuning in the OLMo2-7B-SFT model on CommonsenseQA under identical compression schedules. The direct comparison between these loss functions highlights their critical role in recovery capabilities after pruning, emphasizing loss function design as a pivotal factor for optimizing compressed models in data-sparse scenarios. These insights provide actionable guidance for developing efficient edge AI systems, particularly in resource-constrained environments. Future work will integrate multi-axis pruning strategies with quantization and advanced distillation techniques to further enhance performance, leveraging this foundational understanding to push state-of-the-art results for edge-deployable devices.



\section{Conclusion}
\label{conclusion}
In this paper, we carried out a controlled study to isolate the impact of the recovery loss function under a fixed, MLP-only pruning baseline.  Specifically, we compared Cross-Entropy fine-tuning (\textbf{L2PFT}) against KL-divergence self-distillation (\textbf{L2PSD}) on the OLMo2-7B-SFT model for CommonsenseQA.  Our experiments demonstrate that \textbf{L2PSD} matches or exceeds \textbf{L2PFT}-achieving a 3–5\% improvement in test accuracy at 50\% parameter retention-highlighting the non-trivial role of the loss function in post-prune recovery.  Furthermore, our ablation study confirms that increased training data reduces prediction uncertainty, with self-distillation mitigating entropy reduction through soft-target training.

These findings underscore that, even when using a simple, layer-wise $L_2$-norm pruning on only the MLP blocks, the choice of re-training loss materially affects compressed-model performance.  By demonstrating the regularizing benefits of KL-divergence in a resource-constrained setting, this work contributes a clear, component-level insight into AI model optimization.

\textbf{Future work} will build on these results by extending pruning beyond the MLP layers to the full Transformer architecture and refining sparsity schedules.  Our goal is to push the accuracy–compression frontier further, exploring more comprehensive pruning strategies and advanced re-training protocols to optimize large foundational models for edge deployment.

\section*{Acknowledgments}
This work was supported by the DEVCOM Army Research Laboratory under
Cooperative Agreement No. W911NF2420176.

\bibliographystyle{unsrt}  
\bibliography{references}

\end{document}